\documentclass[letterpaper]{article} 
\usepackage{aaai24}  
\usepackage{times}  
\usepackage{helvet}  
\usepackage{courier}  
\usepackage[hyphens]{url}  
\usepackage{graphicx} 
\usepackage{natbib}  
\usepackage{caption} 
\usepackage{algorithm}
\usepackage{algorithmic}

\usepackage{newfloat}
\usepackage{listings}
\DeclareCaptionStyle{ruled}{labelfont=normalfont,labelsep=colon,strut=off} 
\floatstyle{ruled}
\newfloat{listing}{tb}{lst}{}
\floatname{listing}{Listing}
\usepackage{xcolor}
\usepackage{graphicx}

\usepackage{booktabs}
\usepackage{multirow}
\usepackage{hhline}
\usepackage{subcaption}

\usepackage{tabularx}
\usepackage{arydshln}
\usepackage{pifont}
\usepackage{amsmath}

\title{Response Enhanced Semi-supervised Dialogue Query Generation}
\author{
    Jianheng Huang\textsuperscript{\rm 1,2,3}\equalcontrib,
    Ante Wang\textsuperscript{\rm 1,2,3}\equalcontrib,
    Linfeng Gao\textsuperscript{\rm 1,3},
    Linfeng Song\textsuperscript{\rm 4},
    Jinsong Su\textsuperscript{\rm 1,2,3}\thanks{Corresponding author.}
}
\affiliations{
    \textsuperscript{1}School of Informatics, Xiamen University, China\\ 
    \textsuperscript{2}Shanghai Artificial Intelligence Laboratory, China\\ 
    \textsuperscript{3}Key Laboratory of Digital Protection and Intelligent Processing of Intangible Cultural Heritage \\
    of Fujian and Taiwan (Xiamen University), Ministry of Culture and Tourism, China\\
    \textsuperscript{4}Tencent AI Lab

    \{enatsu,\ wangante,\ 22920202200799\}@stu.xmu.edu.cn,\\ lfsong@tencent.com,\ jssu@xmu.edu.cn
}

\usepackage{bibentry}

\begin{document}

\maketitle

\begin{abstract}
Leveraging vast and continually updated knowledge from the Internet has been considered an important ability for a dialogue system.
Therefore, the dialogue query generation task is proposed for generating search queries from dialogue histories, which will be submitted to a search engine for retrieving relevant websites on the Internet.
In this regard, previous efforts were devoted to collecting conversations with annotated queries and training a query producer (QP) via standard supervised learning.
However, these studies still face the challenges of data scarcity and domain adaptation.
To address these issues, in this paper, we propose a semi-supervised learning framework -- SemiDQG, to improve model performance with unlabeled conversations. 
Based on the observation that the search query is typically related to the topic of dialogue response, we train a response-augmented query producer (RA) to provide rich and effective training signals for QP.
We first apply a similarity-based query selection strategy to select high-quality RA-generated pseudo queries, which are used to construct pseudo instances for training QP and RA.
Then, we adopt the REINFORCE algorithm to further enhance QP, with RA-provided rewards as fine-grained training signals.
Experimental results and in-depth analysis of three benchmarks show the effectiveness of our framework in cross-domain and low-resource scenarios. Particularly, SemiDQG significantly surpasses ChatGPT and competitive baselines. Our code is available at \url{https://github.com/DeepLearnXMU/SemiDQG}.
\end{abstract}

\section{Introduction}

Recent years have witnessed the burgeoning of pre-trained language models (PLMs) \cite{lewis2019bart,raffel2020exploring} and large language models (LLMs), which effectively improve the performance of various downstream tasks and pave the way for artificial general intelligence (AGI) \cite{goertzel2007artificial}.
Despite the variation in size, these models can still fail to generate factual content, which is known as \emph{hallucination} \cite{Ji_2023,openai2023gpt4}. To tackle this issue, researchers have explored incorporating external knowledge from search engines \cite{komeili-etal-2022-internet}.
Typically, to bridge a model with a search engine, a query producer is used to generate search queries for retrieving relevant websites.
In this work, we focus on \emph{dialogue query generation}, which is more challenging as it has to mine user intents from complex dialogue contexts. 

To train such a query producer, previous studies resort to supervised learning, where conversations with annotated search queries are used to fine-tune a pre-trained model \cite{lewis2019bart,raffel2020exploring}. However, it is costly to construct a dataset with enough human annotations, and the trained model may still have a disappointing performance in out-of-domain conversations.
A common practice to tackle these issues is semi-supervised learning \cite{yarowsky1995unsupervised,blum1998combining}, which has been widely investigated in both CV \cite{rosenberg2005semi} and NLP \cite{zhang-zong-2016-exploiting,he2020revisiting} fields.
It suits the dialogue query generation task well because abundant conversations without annotated queries are easy to obtain. 
As implemented in self-training, we expect the model to generate pseudo queries for unlabeled conversations. 
While in practice, some pseudo queries are often unsatisfying, which may lead to error accumulation and model performance degradation. It can be said that the challenge of effectively collecting high-quality pseudo queries to construct pseudo instances continues to be a hurdle in this task.

\begin{table}[t!] \small
    \centering
    \tabcolsep=5pt
    \begin{tabularx}{0.475\textwidth}{r r X}
        \textbf{Example 1} \\
        \midrule
        \multirow{3}*{History}
        & System: & Ever been to \textbf{Ireland} in the \underline{North Atlantic}? Heard it is lovely. \\
        & User: & I have not been there but I'd love to \\
        \hdashline
        \multirow{3}*{Response}
        & System: & \textbf{It}'s not too big but \textbf{it} is the third largest island in Europe so not too small, like a lively and nice place.\\
        \midrule
        Gold query & \multicolumn{2}{l}{\textit{ireland}} \\
        QP's prediction & \multicolumn{2}{l}{\textit{north atlantic} \ding{56}} \\
        RA's prediction & \multicolumn{2}{l}{\textit{ireland} \ding{52}} \\
        \midrule
        \textbf{Example 2} \\
        \midrule
        \multirow{9}*{History}
        & User: & I love to go \textbf{bowling} with my family, but I'm a horrible bowler.  Do you like it? \\
        & System: & Oh, yes, I love \textbf{bowling}.  Rolling balls down the lane and knocking down the pins gives me a charge. \\
        & User: & I know!  I love it when I just knock one down - lol!! My kids want to win, I just like playing. \\
        \hdashline
        \multirow{3}*{Response} 
        & System: & Since \textbf{it} is one of the major throwing sports, \textbf{it} is a lot like the \underline{javelin throw}. \\
        \midrule
        Gold query & \multicolumn{2}{l}{\textit{bowling}} \\
        QP's prediction & \multicolumn{2}{l}{\textit{bowling} \ding{52}} \\
        RA's prediction & \multicolumn{2}{l}{\textit{javelin throw} \ding{56}} \\
    \bottomrule
    \end{tabularx}
    \caption{Two examples from Wizard-of-Wikipedia (WoW, \citealt{dinan2018wizard}) with corresponding dialogue responses, gold queries, and model predictions. QP and RA denote the standard query producer and the response-augmented model respectively. Here we highlight the main topics or their referring expressions that help predict gold queries in bold and mark the misleading concepts with underlines.}
    \label{tab:example}
\end{table}

Fortunately, we notice that a search query can be highly relevant to the topic of its corresponding dialogue response. 
When augmenting the input with response information, the model can often generate better search queries. 
As illustrated in the first case of Table \ref{tab:example}, the standard query producer (QP) solely incorporates the dialogue history as input and mistakenly recognizes ``\textit{north atlantic}'' as the query. In contrast, the response-augmented query producer (RA) accurately predicts the correct query by inferring the mainly discussed topic ``\textit{ireland}'' (referred to by ``\textit{it}'') in the response. 
This demonstrates the potential of RA to generate high-quality pseudo queries which can subsequently be used to construct pseudo instances for training QP.\footnote{Note that we focus on improving the performance of QP in that the response information is inaccessible in practical application.}
However, we notice that RA may also generate some low-quality queries especially when it is overly influenced by the response. In the second case of Table \ref{tab:example}, RA ignores the principal topic ``\textit{bowling}'' in the history, but mistakenly takes ``\textit{javelin throw}'', another topic in the response, as the prediction. Therefore, it is worth exploring ways to select high-quality RA-generated pseudo queries.

Based on the observations above, we propose a novel framework \textbf{Semi}-supervised \textbf{D}ialogue \textbf{Q}uery \textbf{G}eneration (SemiDQG) which effectively improves QP with the guidance of RA. Specifically, we first train QP and RA on a labeled dataset. Subsequently, we leverage the capabilities of RA to generate pseudo queries for an unlabeled dataset and introduce a query selection strategy based on the prediction similarity between QP and RA to select high-quality RA-generated queries (e.g., ``\textit{ireland}" in Table \ref{tab:example}). In a semi-supervised manner, these selected queries are used to construct pseudo instances, thereby enhancing the performance of both models. Finally, to further enhance QP, we adopt the REINFORCE algorithm \cite{williams1992simple} with RA-provided rewards, serving as fine-grained training signals, based on QP-generated candidate queries. 
Both pseudo instance construction and the reinforcement learning approach proposed above can jointly consider the output features from both QP and RA.
Thus, it can fully utilize the training signals from RA spanning different levels of granularity and effectively alleviate the negative effect stemming from input discrepancy between the two models.

We conduct experiments in cross-domain and low-resource scenarios respectively. In the cross-domain scenario, we construct Wizard-of-Internet (WoI, \citealt{komeili-etal-2022-internet})\,$\rightarrow$\,Wizard-of-Wikipedia (WoW, \citealt{dinan2018wizard}) in English, and DuSinc \cite{zhou2022link}\,$\rightarrow$\,KdConv \cite{zhou2020kdconv} in Chinese. In the low-resource scenario, we focus on WoI as it provides more data for better evaluation.
Experiment results show that SemiDQG significantly outperforms ChatGPT and various baselines. Moreover, in-depth analysis validates the effectiveness of the proposed query selection strategy and reinforcement learning method in our framework.

\begin{figure*}[tbp]
    \centering
    \includegraphics[width=\linewidth]{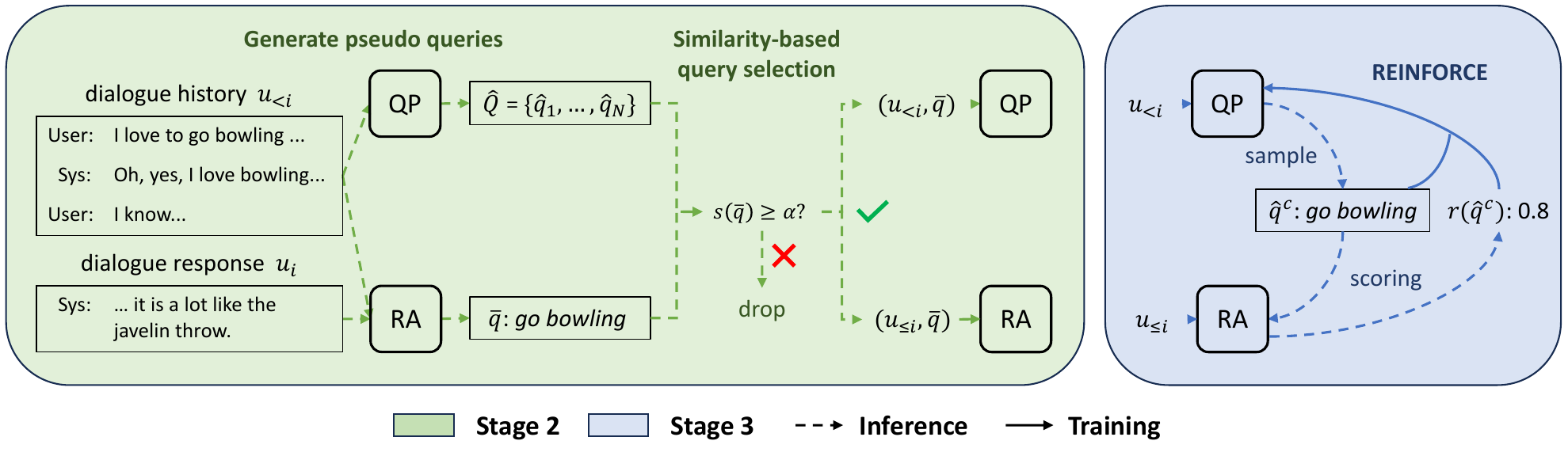}
    \caption{Our proposed Semi-supervised Dialogue Query Generation (SemiDQG) framework. 
    In Stage 1, we train QP and RA via standard supervised training on labeled data (not shown for clarity).
    In Stage 2, for each unlabeled conversation, we use RA to generate its pseudo queries $\bar{q}$. We only keep the query whose similarity score $s(\bar{q})$ exceeds a given threshold $\alpha$ to construct a pseudo instance. We use these high-quality pseudo instances to train QP and RA.
    In Stage 3, QP is further enhanced using RA-guided reinforcement learning.
    }
    \label{fig:arch}
\end{figure*}

\section{Related Work}
\paragraph{Search Query Generation}
Using a search engine to exploit knowledge from the Internet is gaining popularity for benefiting various knowledge-intensive tasks, such as open-domain QA \cite{qi-etal-2019-answering,nakano2022webgpt}, and dialogue response generation \cite{komeili-etal-2022-internet, glaese2022improving}. 
Early attempts simply take user questions or keywords as search queries but have been proven to be ineffective when handling distinct domains \cite{XIE2023103382} or complex dialogue contexts \cite{Wang_2023}.
Recent work \cite{komeili-etal-2022-internet, zhou2022link, Wang_2023} usually trains a query producer to extract or generate search queries, with query generation more popular due to the limitation of extraction.
With the release of various query generation datasets \cite{komeili-etal-2022-internet, zhou2022link}, researchers can build their query producers in supervised learning manners. 
As query annotations are costly to collect, some researchers \cite{qi-etal-2019-answering, Wang_2023, wang-etal-2023-domain} introduce additional supervision signals to train their query producers. 

Very recently, many LLM products \cite{thoppilan2022lamda, glaese2022improving} use prompting techniques to generate search queries instead of adopting an independent query producer.
However, prompting techniques heavily rely on the ability of LLMs to understand the prompt. After a comparison of these two strategies, our experimental results show that even ChatGPT still shows inferior performance than a smaller task-specific model.

\paragraph{Semi-supervised Learning} 
As a branch of machine learning, semi-supervised learning exploits the knowledge from unlabeled data when labeled data is limited. 
In this regard, typical methods mainly include self-training \cite{yarowsky1995unsupervised}, co-training \cite{blum1998combining, zhou2004democratic}, tri-training \cite{zhou2005tri}, and so on.
Among them, self-training is one of the earliest approaches and continues to gain popularity in recent years \cite{amini2022self}. 
For a specific task, it improves a model by iteratively enriching the training data with selected pseudo instances.
In NLP fields, several studies have investigated self-training on text generation tasks, such as neural machine translation \cite{he2020revisiting}, text summarization \cite{he2020revisiting}, and question generation \cite{kulshreshtha-etal-2021-back}.
Nevertheless, it is challenging to collect appropriate pseudo instances, potentially hindering the progress in building more powerful models.
In this work, we focus on leveraging semi-supervised learning to further enhance the query producer, as illustrated in the following section.

\section{Our Framework}

Figure \ref{fig:arch} illustrates the procedure of our proposed SemiDQG, which can be roughly separated into three stages.
In \textbf{Stage 1}, we train a standard query producer (QP) and a response-augmented query producer (RA) on a labeled dataset via supervised learning.
In \textbf{Stage 2}, both QP and RA generate pseudo queries for an unlabeled dialogue corpus. 
Then, based on the prediction similarity between RA and QP, we select high-quality RA-generated queries to construct pseudo instances for training these two models.
Nevertheless, due to the discrepancy between QP and RA, these pseudo instances might not effectively guide QP. Thus, in \textbf{Stage 3}, we employ reinforcement learning to further improve QP with RA providing rewards as fine-grained training signals.
Detailed descriptions will be provided in the following subsections.

\subsection{Stage 1: Train QP and RA with Supervised Learning}

As described above, under our framework, we train a QP and an RA via supervised learning in this stage. Formally, given the dialogue history $u_{<i}$\ =\ $u_1,...,u_{i-1}$, both QP and RA aim to predict the target query $q$. The difference between QP and RA lies in that RA takes the dialogue response $u_i$ as additional input, which is inaccessible in practical application. 

Following the previous study \cite{komeili-etal-2022-internet}, we choose pre-trained T5 \cite{raffel2020exploring} as the basic model for QP and RA, and fine-tune them on conversations with annotated queries. 
For each instance, we take the cross-entropy loss (CE) as the training objective:
\begin{equation}
\begin{aligned}
\mathcal{L}_{qp} &= - \log p(q \mid u_{<i}; \theta_\mathrm{qp}), \\
\mathcal{L}_{ra} &= - \log p(q \mid u_{\leq i}; \theta_\mathrm{ra}),   
\end{aligned}
\label{eq:ce_loss}
\end{equation}
where $\theta_\mathrm{qp}$ and $\theta_\mathrm{ra}$ denote the parameters of QP and RA respectively. 

\subsection{Stage 2: Semi-supervised Learning with Similarity-based Query Selection}

Once the above training is completed, we use RA to generate queries for an unlabeled dialogue corpus and select high-quality queries to construct pseudo instances, which are finally used to enhance QP and RA.
Please note that, unlike the standard self-training, we take advantage of RA rather than QP in generating pseudo queries and constructing instances for QP.

One important step of the above process is the quality evaluation of RA-generated queries. Intuitively, the most direct approach is to use their predictive probabilities as the evaluation metric.
However, modern neural networks are often poorly calibrated \cite{guo2017calibration} and their predictive probabilities may not be reliable.
To deal with this issue, we also use QP to generate queries for the unlabeled dialogue corpus and then evaluate the quality of RA-generated queries by the prediction similarity between RA and QP.

Formally, given a dialogue history $u_{<i}$ and response $u_i$ in the unlabeled corpus, we use RA to generate a query $\bar{q}$ and adopt QP to generate $N$ queries $\hat{Q}$\ =\ $\{\hat{q}_1,...,\hat{q}_N\}$ with only $u_{<i}$ as input.

Then we quantify the quality of RA-generated query $\bar{q}$ by the following similarity score:
\begin{equation}
\begin{aligned}
s(\bar{q}) &= \max \{\mathcal{F}_{sim} (\bar{q}, \hat{q}_i)\}_{\hat{q}_i \in \hat{Q}},    
\end{aligned}
\end{equation}
where $\mathcal{F}_{sim}(*)$ denotes a text similarity function that returns the score of a specific quantitative metric (e.g., Unigram F1 and ROUGE) or a semantic similarity model such as Sentence-BERT \cite{reimers2019sentencebert}.
Note that if $\bar{q}$ is overly influenced by the response information, it will contain unrelated concepts from the response and thus will have a low similarity score. 

Afterward, we select high-quality RA-generated queries, whose similarity score exceeds a pre-determined threshold $\alpha$, to construct pseudo instances with the corresponding dialogue histories.
Next, we use these pseudo instances to further train QP using the CE loss again (See Equation \ref{eq:ce_loss}).

Particularly, during this process, the training strategies we adopt vary slightly in different scenarios.
Concretely, in the cross-domain scenario, we directly fine-tune the best checkpoints of QP from Stage 1 on RA-labeled pseudo instances, as implemented in \cite{meng-etal-2023-general}.
While in the low-resource scenario, we follow \citet{he2020revisiting} to retrain QP on RA-labeled pseudo instances.
Finally, it should be noted that we also further train RA in the above manners to facilitate the subsequent training.

\subsection{Stage 3: RA-guided Reinforcement Learning}

Unfortunately, during the stage above, there are still some low-quality pseudo instances left, which may have negative effects.
More importantly, QP still fails to fully utilize useful fine-grained training signals from RA by training on pseudo instances only.
Thus, in Stage 3, we adopt the REINFORCE algorithm \cite{williams1992simple} to tackle these problems. 

Concretely, for each instance in an unlabeled dialogue corpus, we first sample $N_c$ candidate queries from the predictive distribution of QP.
Here we follow \citet{liu-etal-2022-brio} to calculate the length-normalized log probability of QP for each candidate query $\hat{q}^c$:
\begin{equation}
    f_{qp}(\hat{q}^c)=\frac{\sum_{j}\log p(\hat{q}^c_j \mid u_{<i}, \hat{q}^c_{<j}; \theta_\mathrm{qp})}{|\hat{q}^c|},
\label{eq:model_score}
\end{equation}
where $\hat{q}^c_j$ denotes the $j$-th query token. Furthermore, using a \textit{softmax} normalization, we derive a predictive distribution over all candidate queries, acting as the stochastic policy to sample $\hat{q}^c$.

Then we explore the following two kinds of reward $r(\hat{q}^c)$:

\begin{itemize}
    \item \textbf{Prob-based Reward} Similar to Equation \ref{eq:model_score}, we feed each candidate query $\hat{q}^c$ into RA and calculate its length-normalized log probability, denoted as $f_{ra}(\hat{q}^c)$. We directly use this probability as the reward:
    \begin{equation}
    \begin{aligned}
    r(\hat{q}^c)=f_{ra}(\hat{q}^c),
    \end{aligned}
    \end{equation}
    \item \textbf{Rank-based Reward} 
    We sort all candidate queries by $f_{ra}(\hat{q}^c)$ and design the following reward: 
    \begin{equation}
    \begin{aligned}
    r(\hat{q}^c) =&\ \frac{1}{1 + g(\hat{q}^c)},
    \end{aligned}
    \end{equation}
    where $g(*)$ is a function that returns the descending order of input queries according to $f_{ra}(\hat{q}^c)$. 
    Note that We perform normalization across all $N_c$ candidate queries to reduce variance in gradient estimation. This procedure will punish pseudo queries with low rankings.
\end{itemize}

Finally, QP can be trained with the guidance of reward:
\begin{equation}
\mathcal{L}_{rl} = - r(\hat{q}^c) \log p(\hat{q}^c \mid u_{<i}; \theta_\mathrm{qp}).
\end{equation}
Intuitively, the reward provided by RA is a fine-grained training signal compared to the pseudo queries in Stage 2.

\section{Experiments}

\subsection{Setup}


\paragraph{Datasets} We conduct experiments in both cross-domain and low-resource scenarios across three benchmarks. In the cross-domain scenario, we explore Wizard-of-Internet (WoI, \citealp{komeili-etal-2022-internet})\,$\rightarrow$\,Wizard-of-Wikipedia (WoW, \citealp{dinan2018wizard}) in English, and DuSinc \cite{zhou2022link}\,$\rightarrow$\,KdConv \cite{zhou2020kdconv} in Chinese.
In the low-resource scenario, we focus on WoI, as it provides more high-quality query annotation data for better evaluation. 

\begin{itemize}
    \item \textbf{Wizard-of-Internet (WoI)} A comprehensive dataset providing conversations with search query annotations and websites retrieved from the Bing Search API\footnote{\url{https://www.microsoft.com/en-us/bing/apis/bing-web-search-api}}. 
    \item \textbf{Wizard-of-Wikipedia (WoW)} A popular dialogue dataset, with each utterance grounded on a Wikipedia page. We follow \citet{Wang_2023} to use Wikipedia Search\footnote{\url{https://www.wikipedia.org}} as the search engine and evaluate the quality of search queries by comparing retrieved Wikipedia page titles with the gold one.
    \item \textbf{DuSinc} A Chinese open-domain dialogue dataset with annotated search queries. We use its publicly available part\footnote{\url{https://aistudio.baidu.com/aistudio/datasetdetail/139431}} for experiments.
    \item \textbf{KdConv} A Chinese multi-domain knowledge-driven conversation dataset containing knowledge graph (KG) triplets where dialogue responses may need knowledge from a KG. For each triplet, we use the concatenation of the subject and the predicate as the gold query.
\end{itemize}

\paragraph{Evaluation Metrics}

All the metrics we use to evaluate the model performance are listed below:

\begin{itemize}
    \item \normalsize \textbf{Recall-k (R@k)} We use this metric only on WoW. It is decided by the recall of the target Wikipedia page title when feeding the top-$k$ ($k \in \{1,3\}$) predicted queries to Wikipedia search.
    \item \normalsize \textbf{Unigram F1 (Uni. F1)} We use this metric on all the datasets. It measures the unigram overlap between the prediction and gold reference.
    \item \normalsize \textbf{BLEU} It is a typical metric for text generation tasks that mainly focus on the precision of $n$-gram for the prediction against the gold reference. We use \textit{sacrebleu} \cite{post-2018-call} for \textit{BLEU-1/2} calculation.
    \item \normalsize \textbf{ROUGE} As another commonly used evaluation metric for text generation, it accounts for both precision and recall, thus providing more comprehensive scores. We report \textit{ROUGE-1/2/L} using Google's implementation\footnote{\url{https://github.com/google-research/google-research/tree/master/rouge}}.
\end{itemize}

\paragraph{Baselines} We compare our proposed SemiDQG with the following baselines:

\begin{itemize}
    \item \textbf{T5-base} A fine-tuned T5-base model \cite{raffel2020exploring} on a labeled dataset following \citet{komeili-etal-2022-internet}, same as QP in Stage 1 as mentioned above.
    \item \textbf{Self-training(scratch)} A model initialized from the original T5-base parameters and trained on the QP-labeled pseudo instances following \citet{he2020revisiting}.
    \item \textbf{Self-training(QP)} A model initialized from trained QP in Stage 1 and then tuned on \textit{self}-labeled pseudo instances.
    \item \textbf{Self-training(joint)} The original T5-base model fine-tuned on the combination of synthetic data and authentic data following \citet{he2020revisiting}.
    \item \textbf{QP-ext/QP-gen} \cite{Wang_2023} Different types of QPs, based on extraction and generation respectively. Both are trained with cheap noisy supervision, taking feedback from the Wikipedia search as training signals, and significantly surpass unsupervised keyword extraction methods.
    \item \textbf{KD (RA\,$\rightarrow$\,QP)} A model that adopts vanilla knowledge distillation \cite{hinton2015distilling, miao-etal-2023-exploring}, where the student model (QP) is trained to fit predictions of the teacher model (RA).
    \item \textbf{ChatGPT} We utilize the official \texttt{gpt-3.5-turbo} API\footnote{\url{https://openai.com/blog/openai-api}} to perform inference by in-context learning with 3 or 8 demonstrations following \citet{ye2023incontext}.
\end{itemize}

\paragraph{Implementation Details}
For all pre-trained models used in this work, We utilize the checkpoints from Huggingface\footnote{\url{https://huggingface.co/models}}, with different T5-base variants according to languages.
For English datasets, we use the \texttt{t5-base}. While for Chinese datasets, we employ the \texttt{Langboat/mengzi-t5-base} \cite{zhang2021mengzi}.
During training, we apply an Adam optimizer, with a linear scheduler and an initial learning rate of 3e-5. We use a batch size of 64 for cross-domain experiments and 16 for low-resource counterparts.  
For the main experiments, we set $N$\ =\ $1$ for query selection, and use Unigram F1 as the default $\mathcal{F}_{sim}$. The selection of hyperparameter $\alpha$ for WoW/WoI/KdConv is 1.0/1.0/0.5, respectively. We set $N_c$\ =\ $10$ for rank-based reward in the cross-domain scenario and $N_c$\ =\ $3$ for other settings. Next, we analyze the selection of some key hyperparameters as follows.

\begin{table}[t]
\centering
\scalebox{1}{
\begin{tabular}{lcc}
\toprule
\multirow{1}{*}{\textbf{Model}}     & \multicolumn{1}{c}{\textbf{WoW}}                        & \multicolumn{1}{c}{\textbf{KdConv}}   \\
\midrule
QP(Stage 1) & 40.68 & 56.85 \\
\hdashline
\textit{Stage 2}\\
~~w/ RA-labeled instances & 27.31 & 63.50 \\
~~~~+ Uni. F1 as $\mathcal{F}_{sim}$ & 42.92 & 64.68 \\
~~~~+ Sentence-BERT as $\mathcal{F}_{sim}$ & 41.34 & 64.11 \\
\bottomrule
\end{tabular}
}
\caption{
Results on development sets of WoW and KdConv with different $\mathcal{F}_{sim}$ in Stage 2.}\label{tab:fsim}
\end{table}

\subsection{Development Results}

\paragraph{Selection of $\mathcal{F}_{sim}$} In Stage 2, we investigate two types of $\mathcal{F}_{sim}$: Unigram F1 as the quantitative metric and Sentence-BERT as the semantic similarity model. 
As observed in Table \ref{tab:fsim}, both $\mathcal{F}_{sim}$ can effectively enhance QP, and the semantic similarity model does not necessarily yield better results than conventional quantitative metrics. 
Thus, we take Uni. F1 as the $\mathcal{F}_{sim}$ for later experiments.

\begin{figure}
    \centering
    \includegraphics[width=0.95\linewidth]{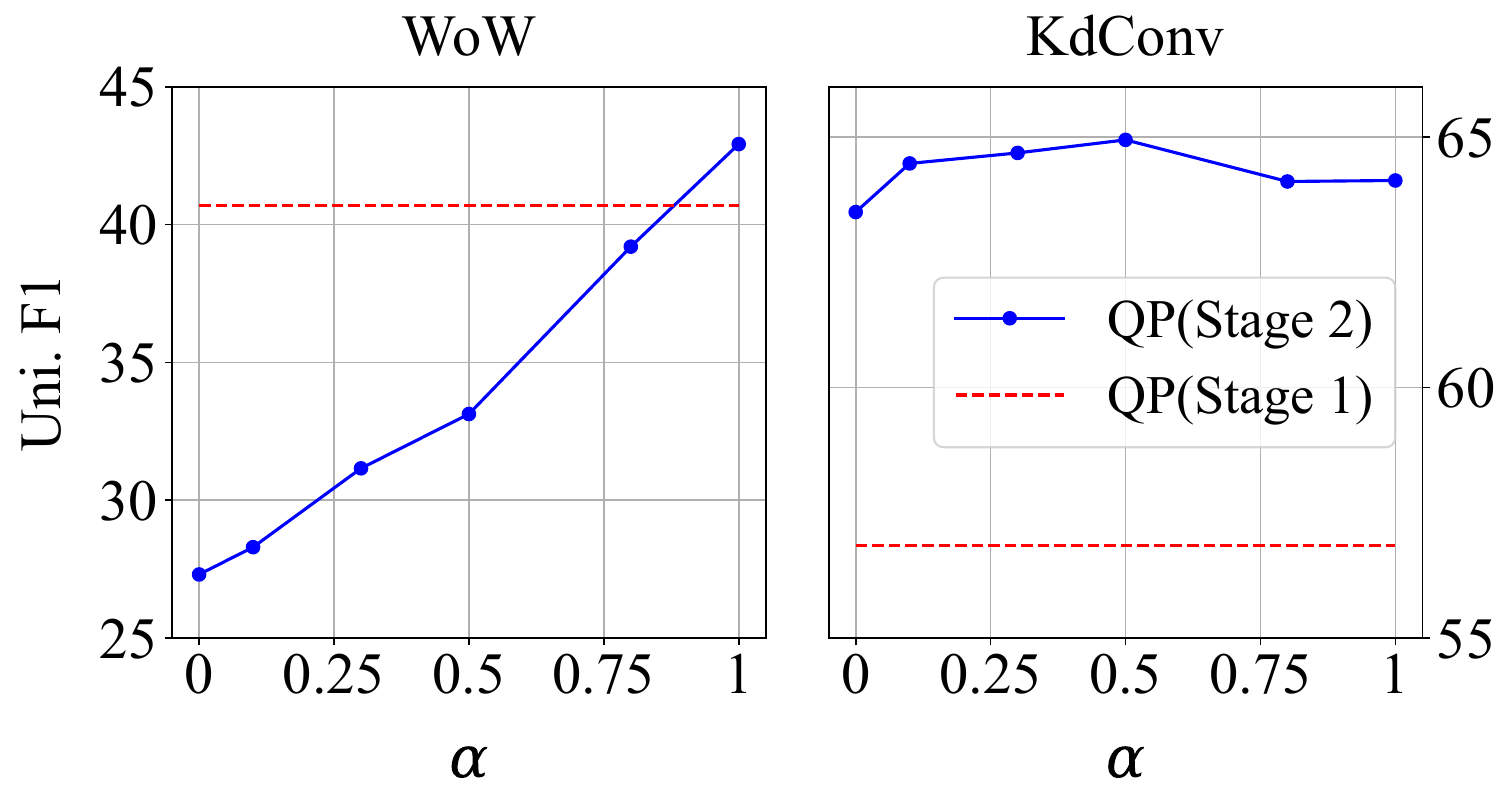}
    \caption{Effect of $\alpha$ on Unigram F1 for development sets of WoW and KdConv in Stage 2.}
    \label{fig:alpha}
\end{figure}

\begin{figure}
    \centering
    \includegraphics[width=0.95\linewidth]{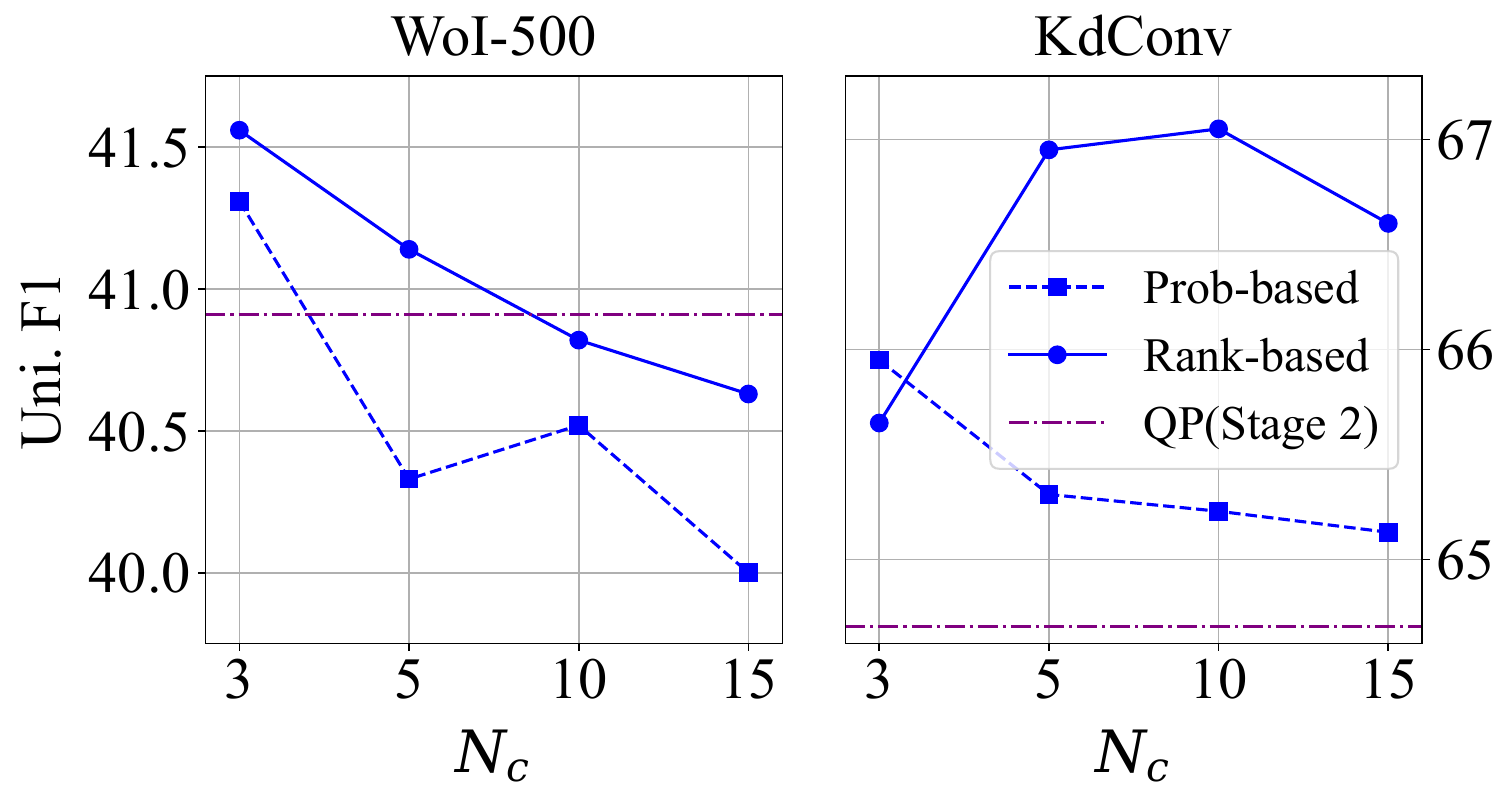}
    \caption{Results on development sets of WoI and KdConv, with different $N_c$ for probability-based and rank-based rewards.}
    \label{fig:reward}
\end{figure}

\paragraph{Selection of $\alpha$} We explore $\alpha$\ =\ $0, 0.1, 0.3, 0.5, 0.8, 1.0$ on WoW and KdConv. As illustrated in Figure \ref{fig:alpha}, the selection of the threshold $\alpha$ significantly affects the model performance.
The model reaches the best performance when $\alpha$\ =\ $1.0 / 0.5$, demonstrating the effectiveness of our similarity-based query selection.
Especially, the model performs even worse than QP(Stage 1) when taking a small $\alpha$ on WoW. This may be attributed to the larger domain gap existing between WoI and WoW.
Nevertheless, this still emphasizes the necessity of adopting our query selection strategy. 
For DuSinc\,$\rightarrow$\,KdConv, the gap may be closer, thus setting a relatively lower $\alpha$ can provide a more diverse set of high-quality pseudo instances to boost the model performance.

\begin{table*}[ht]
\centering
\begin{tabular}{lcccccc}
\toprule
\multirow{2}{*}{\textbf{Model}}     & \multicolumn{3}{c}{\textbf{WoW}}                        & \multicolumn{3}{c}{\textbf{KdConv}}   \\
\cmidrule(lr){2-4} \cmidrule(lr){5-7} 
        & \textbf{Uni. F1} & \textbf{R@1} & \textbf{R@3} & \textbf{Uni. F1}  & \textbf{BLEU-1/2} & \textbf{ROUGE-1/2/L} \\
\midrule
T5-base & 48.11 & 54.30 & 68.59 & 57.13 & 50.86 / 48.37 & 56.91 / 50.74 / 56.72  \\
\midrule
QP-ext$\dagger$ & - & 62.41 & 72.91 & - & - & - \\
QP-gen$\dagger$ & - & 56.77 & 66.08 & - & - & - \\
ChatGPT(3-shot) & 37.88 & 42.63 & - & 49.21 & 39.82 / 36.11 & 49.11 / 41.49 / 48.90   \\
ChatGPT(8-shot) & 41.38 & 46.73 & - & 49.69 & 38.00 / 34.11 & 49.46 / 41.24 / 49.20 \\
\hline
Self-training(scratch) & 37.94 & 33.35 & 45.30 & 58.81 & 53.90 / 51.02 & 58.26 / 52.10 / 58.08 \\
Self-training(QP) & 37.83 & 35.46 & 46.91 & 58.43 & 53.77 / 50.84 & 58.10 / 51.76 / 57.91 \\
Self-training(joint) & 38.61 & 33.80 & 47.16 & 60.10 & 53.71 / 51.11 & 59.79 / 53.78 / 59.64 \\
\hdashline
KD (RA\,$\rightarrow$\,QP)  & 38.42 & 40.35 & 48.31 & 65.60 & 61.88 / 58.88 & 65.40 / 58.95 / 65.32 \\
SemiDQG & \textbf{57.12} & \textbf{63.50} & \textbf{74.89} & \textbf{67.21} & \textbf{62.89} / \textbf{60.19} & \textbf{67.33} / \textbf{60.62} / \textbf{67.30} \\
\bottomrule
\end{tabular}
\caption{Test results on WoW and KdConv in the cross-domain scenario. $\dagger$ denotes the results reported in \cite{Wang_2023}. Note that we only request ChatGPT to generate the most relevant query for each instance, so its R@3 is not applicable.}
\label{tab:cross-domain}
\end{table*}

\paragraph{Selection of $N_c$ and Reward Types} 
We explore the selection of reward types (prob-based and rank-based) with $N_c$\ =\ $3, 5, 10, 15$ for two scenarios separately, as shown in Figure \ref{fig:reward}. 
Generally, prob-based reward only works better when $N_c$ is small, and is inferior to rank-based reward, especially in the cross-domain scenario. This is because poorly calibrated RA cannot provide reasonable confidence scores due to domain discrepancy. 
Furthermore, larger $N_c$ leads to performance degradation in both scenarios since a large $N_c$ will introduce more diverse but low-quality candidates.

\subsection{Main Results}

\paragraph{Cross-domain Scenario}

Table \ref{tab:cross-domain} shows the main results in the cross-domain scenario. 
Overall, SemiDQG achieved the best result, exhibiting remarkable superiority over all baselines across all metrics. While exceeding the typical self-training, it also surpasses other competitive baselines, even the famous LLM product ChatGPT. 
After an in-depth analysis, we have the following conclusions:

(1) Currently accepted LLMs still fail to handle the dialogue query generation task well, despite the application of in-context learning. 
As the number of demonstrations increases from 3 to 8, ChatGPT exhibits some performance improvement on WoW, yet it still falls short of expectations compared to a task-specific model.  
We believe that the capabilities of LLMs should be further explored, as the performance of in-context learning may be constrained.

(2) The two competitive baselines, QP-ext and QP-gen, exhibit performance closest to SemiDQG on WoW. 
However, their training costs are higher due to the use of search engines as feedback. 
Besides, both QP-ext and QP-gen are trained to predict continuous entity spans from inputs. This also makes their approaches impractical on distinct datasets.

(3) Traditional self-training may hurt model performance. As shown in Table \ref{tab:cross-domain}, none of the three self-training variants improve the performance of QP on WoW, and even lead to a decline. Meanwhile, the performance improvement on KdConv is also limited. These results reflect the negative impact of low-quality pseudo instances. 
We also observe that Self-training(scratch) slightly outperforms Self-training(QP) due to different model initializations, aligning with \citet{he2020revisiting}'s findings.

(4) With the guidance of RA, KD (RA\,$\rightarrow$\,QP) beats all self-training approaches on KdConv, demonstrating the necessity of leveraging response information. However, it also performs worse on WoW compared with T5-base similar to self-training baselines. Our SemiDQG successfully improves results on both datasets and significantly outperforms KD (RA\,$\rightarrow$\,QP), validating its effectiveness.

\paragraph{Low-resource Scenario}

\begin{figure}
    \centering    \includegraphics[width=\linewidth]{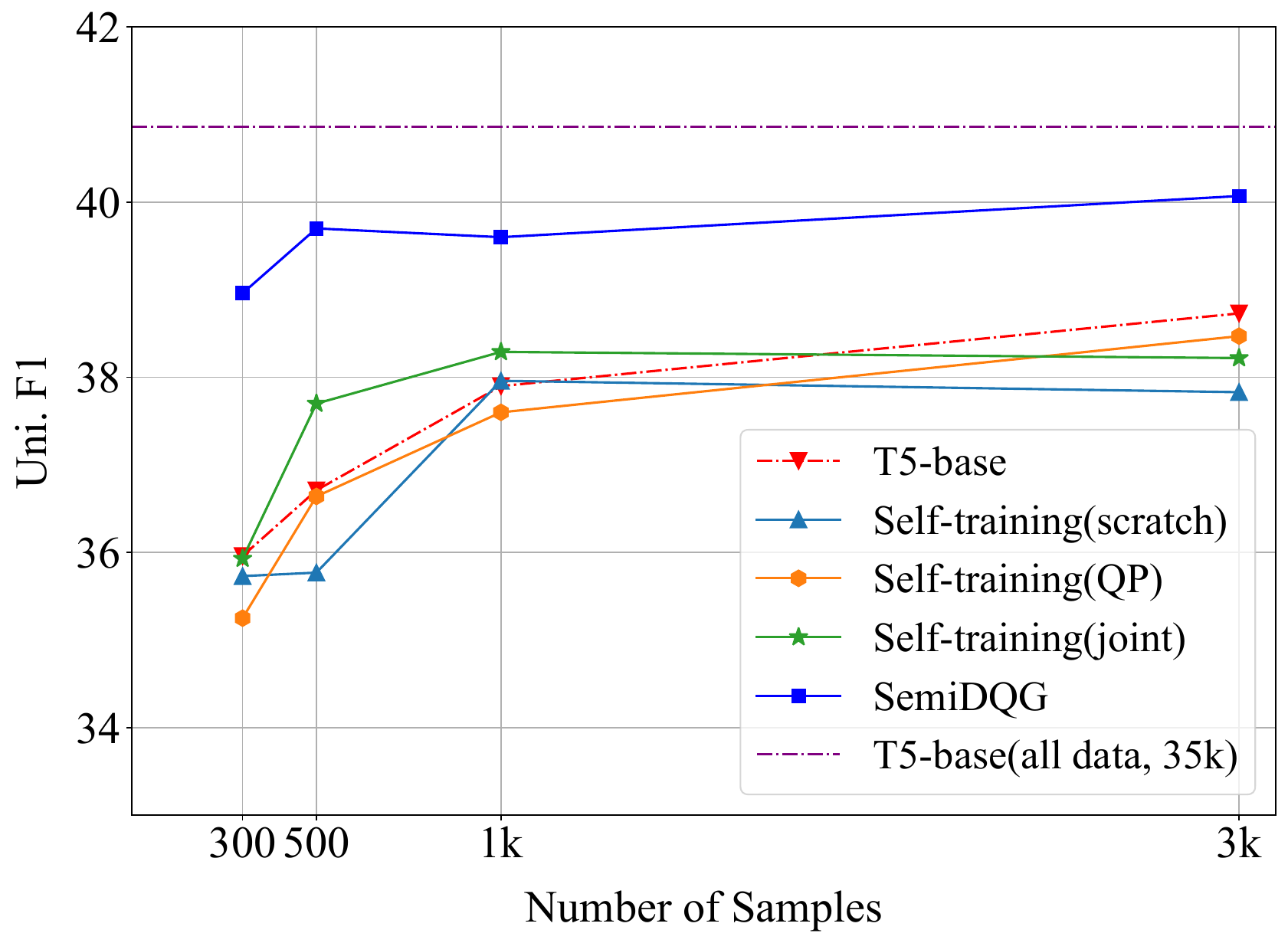}
    \caption{Unigram F1 test results on WoI in the low-resource scenario.}
    \label{fig:low-resouce-woi}
\end{figure}

Figure \ref{fig:low-resouce-woi} depicts that SemiDQG also demonstrates its effectiveness in the low-resource scenario on WoI, which achieves greater performance improvement under extremely low-resource settings (300/500-shot). 
Besides, when using 300 labeled instances, SemiDQG outperforms a T5-base trained with 3k instances, which is 10 times data efficiency.
In addition, similar to the cross-domain results, the performance of the three traditional self-training variants is suboptimal in the low-resource scenario on WoI. This also highlights the limitations of traditional methods and the effectiveness of SemiDQG.

\subsection{Analysis}

In this subsection, we take DuSinc\,$\rightarrow$\,KdConv as an example to conduct a detailed analysis of our proposed framework.

\begin{table}[t]
\centering
\tabcolsep=2pt
\scalebox{0.9}{
\begin{tabular}{lcccccc}
\toprule
\multirow{1}{*}{\textbf{Model}}
 & \textbf{Uni. F1}  & \textbf{BLEU-1/2} & \textbf{ROUGE-1/2/L} \\
\midrule
QP(Stage 1) & 57.13 & 50.86 / 48.37 & 56.91 / 50.74 / 56.72  \\
\hdashline
\textit{Stage 2} \\
~~~~w/ QP-labeled inst. & 58.43 & 53.77 / 50.84 & 58.10 / 51.76 / 57.91 \\
~~~~~~~~$+$ QP-prob sel.  & 59.10 & 53.15 / 50.75 & 58.92 / 52.86 / 58.78 \\
~~~~w/ RA-labeled inst. & 65.60 & \textbf{61.88 / 58.88} & 65.40 / 58.95 / 65.32 \\
~~~~~~~~$+$ RA-prob sel. & 65.73 & 60.06 / 57.69 & 65.71 / 59.60 / 65.62 \\
~~~~~~~~$+$ QP-prob sel.  & 65.01 & 58.81 / 56.55 & 65.04 / 58.91 / 64.97  \\
~~~~~~~~$+$ sim. sel.  &\textbf{66.06} & 61.50 / 58.77 & \textbf{66.12 / 59.67 / 66.07} \\
\hdashline
\textit{Stage 3} \\
~~~~w/ prob-based RL & 66.39 & 62.19 / 59.42 & 66.37 / 59.78 / 66.35 \\
~~~~w/ rank-based RL & \textbf{67.21} & \textbf{62.89 / 60.19} & \textbf{67.33 / 60.62 / 67.30} \\
\hdashline
QP(Stage 1) w/ RL & 64.39 & 59.35 / 56.45 & 64.14 / 57.39 / 64.08 \\
\bottomrule
\end{tabular}
}

\caption{
Ablation studies of QP on the KdConv test set. Here ``instances'', ``similarity'' and ``selection'' are abbreviated as ``inst.'', ``sim.'' and ``sel.'', respectively.
}
\label{tab:ablation}
\end{table}

\begin{table}[t]
\centering
\tabcolsep=2pt
\scalebox{0.9}{
\begin{tabular}{lccc}
\toprule
\multirow{1}{*}{\textbf{Model}}
 & \textbf{Uni. F1}  & \textbf{BLEU-1/2} & \textbf{ROUGE-1/2/L} \\
\midrule
RA(Stage 1) & 64.64 & 55.53 / 53.17 & 64.56 / 58.02 / 64.49 \\
\hdashline
\textit{Stage 2} \\
~~~~w/ RA-labeled inst. & 68.64 & 62.73 / 59.99 & 68.51 / 62.11 / 68.47 \\
~~~~~~~~$+$ RA-prob sel. &  68.37 & 62.81 / 59.98 & 68.33 / 61.72 / 68.26 \\
~~~~~~~~$+$ QP-prob sel. &  \textbf{69.11} & 62.68 / 60.16 & \textbf{69.16 / 62.74 / 69.07} \\
~~~~~~~~$+$ sim. sel. & 68.67 & \textbf{63.77} / \textbf{60.74} & 68.67 / 61.84 / 68.65 \\
\bottomrule
\end{tabular}
}
\caption{Test results of RA variants on KdConv.}
\label{tab:ra}
\end{table}

\paragraph{Similarity-based Query Selection (Stage 2)} We conduct ablation studies as shown in Tables \ref{tab:ablation} and \ref{tab:ra}, comparing our method with query selection based on predictive probabilities of either QP or RA. The main conclusions are as follows:

(1) RA-labeled instances benefit QP more. The utilization of QP-labeled pseudo instances can only slightly enhance QP on KdConv, and the improvement of adopting query selection based on its predictive probability is also limited.

(2) The quality of RA-labeled pseudo instances significantly affects the performance of QP. 
Similarity-based query selection works the best among these selection strategies on KdConv, despite a slight decrease in BLEU-1/2 compared to the vanilla knowledge distillation setting. 
Besides, both QP and RA have difficulty recognizing better pseudo queries, making probability-based query selection less effective than that of the similarity-based counterpart.

(3) RA can also benefit from QP. 
As depicted in Table \ref{tab:ra}, it is challenging for RA to identify instances that can result in significant self-improvement, highlighting its limitation in self-calibration. 
Nevertheless, with the guidance of QP, in terms of either predictive probability or prediction similarity, RA can be further enhanced.

\paragraph{RA as the Reward Model (Stage 3)} Table \ref{tab:ablation} indicates that RA can effectively guide QP to improve model performance, regardless of whether it is adopted directly after Stage 1 or adopted after Stage 2.
As the reward model, RA can provide fine-grained training signals during QP's reinforcement learning process, further tapping into the potential of RA. This validates the necessity and effectiveness of Stage 3. 

We conduct further analysis to demonstrate that RA can provide more reasonable rewards for QP training, which intuitively decides the performance of QP after Stage 3. 
As RA is asked to assess each query $\hat{q}^c$ from the $N_c$ candidates at this stage, we check whether RA can provide a better ranking to these candidate queries according to their quality.

In detail, the following rankings are compared in Table \ref{tab:ra-top1-rank}:
(1) \textbf{QP ranking}. As previously mentioned, we utilize QP to sample the $N_c$ candidate queries via beam search, which naturally results in a descending ranking based on its predictive probability. 
(2) \textbf{RA ranking}. We obtain the ranking by sorting the length-normalized log probability of RA $f_{ra}(\hat{q}^c)$ (See Equation \ref{eq:model_score}) in descending order for each candidate query $\hat{q}^c$.
(3) \textbf{Gold ranking}. We compute the Unigram F1 scores between each $\hat{q}^c$ and the gold query $q$, obtaining an oracle ranking by sorting the scores. 

To evaluate the quality of each ranking, we calculate \textbf{Pearson correlation} coefficients between the QP/RA ranking and the gold ranking and \textbf{Uni. F1 (top-1)}, which gives the Unigram F1 score between the candidate query ranked highest and the gold reference $q$.

As shown in Table \ref{tab:ra-top1-rank}, the RA ranking has a stronger correlation with the gold ranking and gives higher Uni. F1 (top-1) score. This demonstrates the effectiveness of the RA ranking, as it succeeds in allowing high-quality candidate queries to be ranked higher, thus providing more reasonable rewards when applying reinforcement learning.
However, we also notice that there is still a significant performance gap between the RA ranking and the gold ranking. 
We believe that the potential of RA can be further explored.

\begin{table}[]
\centering
\scalebox{0.9}{
\begin{tabular}{lcc}
\toprule
\textbf{Model} & \textbf{Pearson} & \textbf{Uni. F1 (top-1)} \\
\midrule
QP ranking    & 0.3660 & 66.17 \\
RA ranking    & 0.4109 & 67.46 \\
Gold ranking  & 1.0000 & 82.92 \\
\bottomrule
\end{tabular}
}
\caption{The effect of different ranking methods for Pearson correlation coefficient
and top-1 candidate query performance on the KdConv training set in Stage 3.}
\label{tab:ra-top1-rank}
\end{table}

\section{Conclusion}

In this paper, we propose a semi-supervised learning framework, SemiDQG, to enhance the query producer (QP) with the guidance of the response-augmented query producer (RA). 
Taking the dialogue response as an additional feature, RA can provide better training signals for QP training. However, we notice that the input discrepancy between QP and RA will stop our model from further improving.
To alleviate the negative impact of this discrepancy, we jointly consider the output features from both QP and RA as training signals for QP training. 
Specifically, we first apply similarity-based query selection to select high-quality RA-generated pseudo queries for training these models and then adopt RA-guided reinforcement learning to exploit fine-grained knowledge from RA to further improve QP.
Experimental results and in-depth analysis in cross-domain and low-resource scenarios demonstrate the effectiveness of our SemiDQG.

\section{Acknowledgements}

The project was supported by National Key R\&D Program of China (No. 2022ZD0160501), National Natural Science Foundation of China (No. 62276219), and Natural Science Foundation of Fujian Province of China (No. 2020J06001). We also thank the reviewers for their insightful comments.

\bibliography{aaai24}

\end{document}